\documentclass[letterpaper,10pt,conference]{ieeeconf}
\IEEEoverridecommandlockouts
\usepackage{cite}
\usepackage{amsmath,amssymb,amsfonts}
\usepackage{algorithmic}
\usepackage{graphicx}
\usepackage{xurl}
\usepackage{textcomp}
\usepackage{xcolor}
\usepackage{booktabs}
\usepackage{multirow}
\graphicspath{{figures/}}

\makeatletter
\def\bstctlcite{\@ifnextchar[{\@bstctlcite}{\@bstctlcite[@auxout]}}
\def\@bstctlcite[#1]#2{\@bsphack
  \@for\@citeb:=#2\do{%
    \edef\@citeb{\expandafter\@firstofone\@citeb}%
    \if@filesw\immediate\write\csname #1\endcsname{\string\citation{\@citeb}}\fi}%
  \@esphack}
\makeatother

\newcommand{\papertodo}[1]{\par\noindent\textbf{TODO:} #1\par}
\newif\ifpaperreferences
\paperreferencestrue

\title{\LARGE \bf EIDA: Execution-Interface Dynamics Adaptation\\
for Real-to-Sim-to-Real Robot Navigation}
\author{%
  Yiwei Qian$^{1,2}$\authorrefmark{1},
  Shanze Wang$^{1,3}$\authorrefmark{1},
  Qingyuan Hu$^{1,4}$,
  Xinming Zhang$^{1,4}$, and
  Wei Zhang$^{1}$\authorrefmark{2}
  \thanks{*The first two authors contributed equally to this work.}
  \thanks{$^1$Yiwei Qian, Shanze Wang, Qingyuan Hu, Xinming Zhang and Wei Zhang are with Eastern Institute of Technology, Ningbo, China, \texttt{\{szwang,qyhu,xmzhang,zhw\}@eitech.edu.cn}.}
  \thanks{$^2$Yiwei Qian is with the National University of Singapore, Singapore, \texttt{qianyiwei@u.nus.edu}.}
  \thanks{$^3$Shanze Wang is with the Department of Aeronautical and Aviation Engineering, The Hong Kong Polytechnic University, Hong Kong, \texttt{shanze.wang@connect.polyu.hk}.}
  \thanks{$^4$Qingyuan Hu and Xinming Zhang are with the University of Science and Technology of China, Hefei, China, \texttt{\{huqingyuan,xm\_zhang\}@mail.ustc.edu.cn}.}
  \thanks{$^\dagger$Wei Zhang is the corresponding author.}
}

\begin{document}

\bstctlcite{IEEEexample:BSTcontrol}

\maketitle
\thispagestyle{empty}
\pagestyle{empty}
\suppressfloats[t]

\begin{abstract}
Simulation-to-robot transfer can fail when velocity commands produce motion and feedback that differ from those modeled during policy training. We present execution-interface dynamics adaptation (EIDA), which fits these responses from target-platform execution data without reconstructing actuator dynamics. A model of body-frame pose increments updates simulator geometry, while a separate model predicts the velocity feedback observed by the policy; a short history of velocity feedback is included in the policy input. The fitted models are used within a lightweight GPU-parallel simulator. On the full Jackal and Go2 validation sets, the fitted models reduced position and yaw prediction errors relative to the simulator's predefined motion model. Across 100 benchmark navigation environments evaluated in a separate physics-based simulator, EIDA achieved the highest success rate and navigation score among the compared learned policies, both with and without global guidance. Feedback ablations further supported the need to match policy-facing velocity estimates. On a physical Unitree Go2, EIDA reached the goal without collision in all 20 static-scene trials, compared with 4 of 20 for the baseline. These results show that execution-interface adaptation can improve navigation transfer without detailed actuator simulation.

\end{abstract}


\section{Introduction}
\label{sec:introduction}

Efficient and reliable autonomous navigation is fundamental to the safe
operation of mobile robots in complex environments. Deep reinforcement
learning (DRL) can acquire behaviors for collision avoidance and goal reaching
through large-scale interaction~\cite{tai2017navigation,zhang2022ipaprec}, while
GPU-parallel simulation substantially reduces navigation policy training
time~\cite{wang2026flashnav}. However, strong performance in simulation does
not ensure effective transfer to a physical robot.

\begin{figure}[t]
 \centering
 \includegraphics[width=\columnwidth]{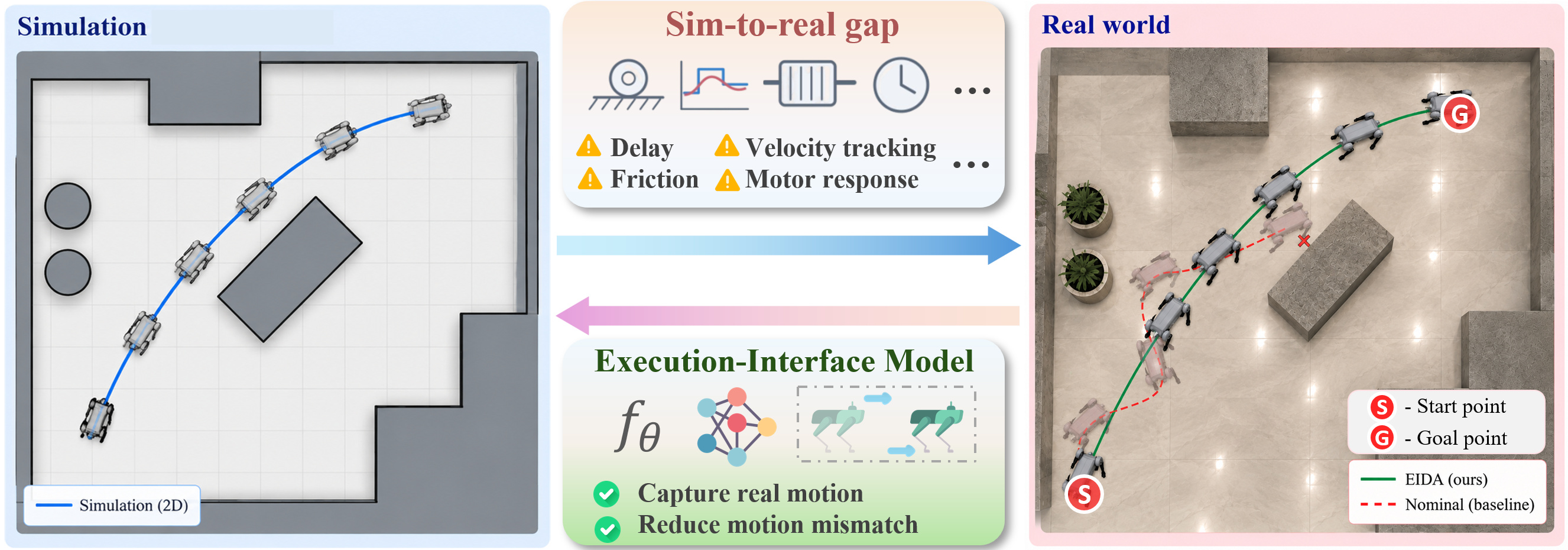}
 \caption{Sim-to-real execution mismatch can cause navigation failures. EIDA
 incorporates models learned from real executions into lightweight simulation
 to improve navigation policy transfer to physical robots.}
 \label{fig:intro-overview}
\end{figure}

This transfer gap involves both perception and execution. Prior studies improve
robustness to visual appearance and sensing errors~\cite{sadeghi2017cad2rl,
loquercio2021wild}, but action modeling and collision handling also affect how
well simulation predicts physical navigation performance~\cite{kadian2020predictivity}.
Velocity commands pass through a platform-specific low-level controller, with
the resulting motion and feedback affected by actuator response, control
latency, and recent commands. Identical commands can therefore produce
different trajectories in simulation and physical execution, potentially
causing drift, turning overshoot, or collisions. At higher speeds, less time is
available for closed-loop correction, making small response errors more
consequential.

Robustness to dynamics mismatch can be improved by training across varied
simulation parameters and adapting their distribution with real
experience~\cite{peng2018dynamicsrandomization,chebotar2019simopt}. Reproducing
detailed actuator and contact dynamics is another option, but increases
computational cost without ensuring better transfer. Truong et
al.~\cite{truong2023rethinking} showed that simplified motion models fitted from
real execution can outperform slower, more detailed simulation in navigation
transfer. In local navigation, collision avoidance and goal progress depend on
pose evolution relative to the environment. This motivates modeling that
evolution together with the velocity feedback available to the policy.
Can navigation policies trained in a lightweight simulator transfer to a
physical robot without reconstructing detailed low-level actuator dynamics?

This paper addresses this question through execution-interface dynamics
adaptation (EIDA), as illustrated in Fig.~\ref{fig:intro-overview}. Using target-system execution data, EIDA fits body-frame pose
increments to update simulator geometry and a separate causal model to predict
the velocity feedback available to the policy. Recent velocity observations
help the policy account for history-dependent execution responses. At
deployment, physical motion and sensor feedback replace the learned simulation
interface, while the existing low-level controller is retained.
EIDA shows that this is possible when pose evolution and policy-facing velocity
feedback are fitted to target-system execution data and incorporated into a
lightweight simulator. Physical Unitree Go2 trials provide direct evidence of
transfer, while evaluation on 100 BARN environments~\cite{perille2020barn}
assesses navigation transfer across simulators. The main contributions of this work are
summarized as follows:
\begin{itemize}
 \item An execution-interface dynamics adaptation framework that learns
 body-frame pose increments from execution data collected on the target system
 and uses them to update simulator geometry during navigation policy training.

 \item A separated representation of motion evolution and velocity feedback
 observable to the policy, together with recent observations that capture
 temporally dependent execution responses.

 \item Validation across wheeled and legged platforms, demonstrating that
 adapting execution responses relevant to navigation improves navigation
 transfer while preserving efficient parallel policy training.
\end{itemize}

\section{Related Work}
\label{sec:related-work}

One class of methods addresses dynamics mismatch by training policies over a
distribution of simulated systems. Dynamics randomization samples physical
parameters during training~\cite{peng2018dynamicsrandomization}, whereas SimOpt
uses experience from the target system to update the distribution of simulation
parameters~\cite{chebotar2019simopt}. Both approaches improve robustness to
variations in dynamics without explicitly replacing the simulator transition
function with a response model fitted to a specific target platform.

Other methods use real execution data to modify simulated transitions or the
mapping from actions to motion. Grounded Action Transformation combines a
target-domain forward model with a simulator inverse model to transform
actions~\cite{hanna2017gat}. Neural-Augmented Simulation learns a recurrent
model of the discrepancy between simulated and physical trajectories and
applies the predicted correction to the simulator~\cite{golemo2018nas}. For
robot navigation, Truong et al.~\cite{truong2023rethinking} showed that simple
motion models fitted from real data can yield better transfer than slower
simulations based on more detailed physics. Bono et
al.~\cite{bono2024navigate} fitted a second-order model that represents the
response of the robot and its low-level controller, and incorporated this model
into Habitat for navigation policy training. In a broader setting, NeRD
replaces low-level dynamics and contact solvers with robot-centric neural
dynamics; the resulting simulator supports policy learning and can be refined
with physical data~\cite{xu2025nerd}.

Learned dynamics models can also be used during online planning rather than to
define the dynamics of the policy training environment. Forward models have
been combined with sampling-based model predictive control to predict robot
trajectories and collision risk during navigation~\cite{kim2022fdm,roth2025fdm}.
By contrast, the Robotic World Model learns autoregressive transitions from
observation--action histories and optimizes policies using imagined
rollouts~\cite{li2025rwm}. RWM-U extends this model with ensemble-based
uncertainty estimates and uncertainty penalties to support policy optimization
from offline data~\cite{li2025rwmu}.

EIDA focuses on the execution interface used during navigation policy training.
It uses body-frame pose increments fitted from target-system execution data to
update the geometric state of the simulator, while a separate causal model
predicts the velocity feedback available to the policy. EIDA therefore differs
from prior methods that transform actions, replace low-level physical
simulation, use learned dynamics for online planning, or model the full
observation transition through a recurrent world model.

\section{Preliminaries}
\label{sec:preliminaries}
\label{sec:method_interface}

We consider goal-directed local navigation, where a robot must reach a
target while avoiding obstacles. At each navigation step, the policy
receives LiDAR measurements, target distance and bearing, and estimated
velocity. It then issues a planar velocity command to an existing low-level
controller.

Let $q_t=(p_t,\psi_t)$ denote the planar reference pose, where
$p_t\in\mathbb{R}^2$ is position and $\psi_t$ is yaw. Let
$u_t\in\mathbb{R}^m$ denote the velocity command in physical units and
$s_t\in\mathbb{R}^m$ the available body-frame velocity estimate. The
command includes forward and angular velocity, with lateral velocity when
supported. The command $u_t$ specifies requested motion, whereas $s_t$
describes the velocity feedback available to the policy. The transition
duration is denoted by $\Delta t_t$. Let $\ell_t\in\mathbb{R}^{30}$
contain sector-minimum LiDAR ranges and $(d_t,\beta_t)$ denote target
distance and body-relative bearing.

A normalized policy action $a_t\in[-1,1]^m$ maps to physical commands as
\begin{equation}
 u_t=u_{\min}+\tfrac{1}{2}(a_t+\boldsymbol{1})
                         \odot(u_{\max}-u_{\min}),
 \label{eq:method_action}
\end{equation}
where $u_{\min}$ and $u_{\max}$ are command limits, $\boldsymbol{1}$ is
the all-ones vector, and $\odot$ denotes elementwise multiplication.

The motion between adjacent reference poses is expressed in the
interval's starting body frame:
\begin{equation}
 \begin{split}
  \delta p_t &= R(\psi_t)^\top(p_{t+1}-p_t),\\
  \delta\psi_t &= \operatorname{wrap}(\psi_{t+1}-\psi_t),
 \end{split}
 \label{eq:method_target}
\end{equation}
with $\delta_t=[\delta p_t^\top,\delta\psi_t]^\top$. Here $R(\psi)$
is the planar rotation matrix and
$\operatorname{wrap}(a)=\operatorname{atan2}(\sin a,\cos a)$.
The increment describes displacement over $\Delta t_t$, rather than an
instantaneous velocity estimate.

The reward combines goal progress with success and collision outcomes:
\begin{equation}
 r_t=\begin{cases}
 r_c, & \text{collision},\\
 r_s, & \text{goal reached},\\
 c_{\text{p}}(d_t-d_{t+1}), & \text{otherwise},
 \end{cases}
 \label{eq:method_reward}
\end{equation}
where $r_s=10$ rewards reaching the goal, $r_c<0$ penalizes collisions,
and $c_{\text{p}}=2$ scales
the dense reward for progress toward
the goal. Collision takes precedence if both terminal conditions hold;
no additional time penalty is applied. Policy learning uses a discount
factor of $\gamma=0.99$.

\section{Methodology}
\label{sec:method}

\begin{figure*}[t]
 \centering
\includegraphics[width=\textwidth]{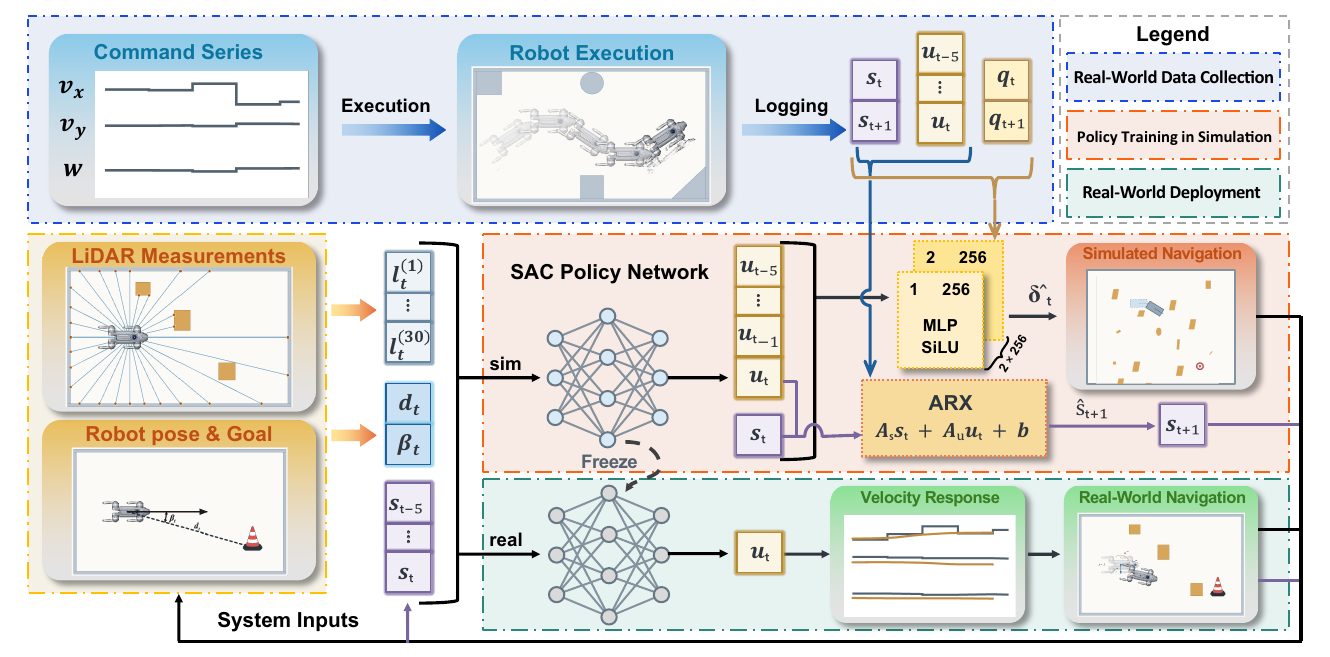}
 \caption{Execution-interface dynamics adaptation (EIDA). Target-system execution data
 identify independent pose-increment and velocity-feedback models. The
 execution-fitted interface supports policy training in simulation; at
 deployment, live sensing and the existing controller replace the fitted
 responses.}
 \label{fig:method_overview}
\end{figure*}

Here, the execution interface denotes the
closed-loop mapping from policy velocity commands to body-frame motion and
velocity feedback, with the existing low-level controller and state
estimator in the loop. Fig.~\ref{fig:method_overview} summarizes the
training and deployment pipeline.

\subsection{Execution-Interface Dynamics Adaptation}
\label{sec:method_models}

To predict motion, the pose model uses recent control inputs.
Its state--command features include the current velocity estimate, the
current command, and $H_u$ preceding commands ordered from most recent
to oldest:
\begin{equation}
 x_t=[s_t^\top,u_t^\top,\ldots,u_{t-H_u}^\top]^\top.
 \label{eq:method_features}
\end{equation}

Reported $(H_u+2)m$ state--command feature dimensions exclude
the scalar interval duration $\Delta t_t$, which is appended for
model evaluation. A pose MLP $f_\theta$ predicts the body-frame pose increment
\begin{equation}
 \widehat\delta_t=f_\theta(x_t,\Delta t_t)
 = [\widehat\delta p_t^\top,\widehat\delta\psi_t]^\top.
 \label{eq:method_mlp}
\end{equation}

Its output represents translation and yaw change expressed in the
interval's starting body frame. Lateral displacement is retained even for
a platform without a lateral command.

This predicted increment directly updates the simulated pose
$\widehat q_t=(\widehat p_t,\widehat\psi_t)$:
\begin{equation}
 \begin{split}
  \widehat p_{t+1}&=\widehat p_t+
                    R(\widehat\psi_t)\widehat\delta p_t,\\
  \widehat\psi_{t+1}&=\operatorname{wrap}
                    (\widehat\psi_t+\widehat\delta\psi_t).
 \end{split}
 \label{eq:method_compose}
\end{equation}

Because the output already represents displacement over $\Delta t_t$, it is
composed once. This update replaces the command-to-motion rule in the navigation
simulator.

While pose increments describe motion over an interval, the deployed
policy receives velocity estimates from a state estimator. Accordingly,
an independent causal autoregressive model with exogenous inputs (ARX) is
fitted to recorded velocity estimates to reproduce this feedback in
simulation. In physical units, its affine form is
\begin{equation}
 \widehat s_{t+1}=g_\phi(s_t,u_t)
 =A_s s_t+A_u u_t+b,
 \label{eq:method_arx}
\end{equation}
where $A_s,A_u\in\mathbb{R}^{m\times m}$ and $b\in\mathbb{R}^m$ are
fitted coefficients that absorb the training normalization. The full
matrices capture cross-axis dependence.

During recursive simulation, both models use the current velocity feedback
and command. Command history and interval duration are additionally provided
to the pose model. Pose prediction advances the physical state, whereas
$\widehat s_{t+1}$ provides the next velocity input to the policy and pose
model. Separate buffers maintain command and observation histories. Ray
casting, relative-target geometry, reward computation, and collision checks
all use the simulated physical pose. This keeps the navigation geometry
consistent throughout the loop. This pose also serves as the
localization proxy. Both models are deterministic; neither localization error
nor predictive uncertainty is explicitly modeled.

\subsection{Model Identification}
\label{sec:method_data}

Manual excitation and policy-shaped trajectories are combined to capture
designed motion responses and navigation-like command changes. Each
sample pairs a recorded command history and available velocity estimate
with adjacent reference poses. The pose target is the increment
$\delta_t$ defined in Eq.~\eqref{eq:method_target}; the observation target
is the next available velocity estimate $s_{t+1}$.

Input features use recorded controller commands and velocity estimates
available at the collection cutoff. Training windows require complete
command histories and remain within continuous valid trajectory segments;
pose gaps are not interpolated.

Data are split by trajectory group, keeping related repeats together.
All normalization statistics and fitted model parameters are obtained
from training data. The observation model $g_\phi$ is first fitted by
weighted ridge regression and then frozen during pose-model training.

Let $w_i$ denote the sample weight determined by data family and
trajectory group, and $\sigma_\delta$ the vector of pose-target standard
deviations. The pose model is trained with the normalized, weighted
single-step loss
\begin{equation}
 \mathcal L_1=
 \frac{\sum_{i\in\mathcal B}w_i\,
 \overline\rho\big((f_\theta(x_i,\Delta t_i)-\delta_i)\oslash\sigma_\delta\big)}
 {\sum_{i\in\mathcal B}w_i},
 \label{eq:method_single_loss}
\end{equation}
where $\mathcal B$ is a minibatch, $\oslash$ denotes elementwise division,
and $\overline\rho(e)=\frac{1}{3}\sum_{j=1}^3\rho(e_j)$. Here $\rho$ is the
Smooth-L1 function: $\rho(a)=a^2/2$ for $|a|<1$ and $|a|-1/2$
otherwise. The weights balance data families and trajectory groups.

Multistep fitting exposes the pose model to the recursive feedback used
in simulation. Each window is initialized with its first recorded
velocity estimate and command history, with predicted and reference poses
at a common local origin. Subsequent steps use recorded commands and
interval durations, while ARX predictions replace subsequent measured
velocity inputs. Predicted and reference increments are composed
separately using Eq.~\eqref{eq:method_compose}.

Let $e_h(W)$ denote the endpoint pose error after $h$ steps in window $W$,
with yaw error wrapped. The multistep objective is
\begin{equation}
 \mathcal L_h=\mathbb E_{W\sim P_h}
 \left[\overline\rho\left(e_h(W)\oslash
       (\sqrt{h}\,\sigma_\delta)\right)\right],
 \label{eq:method_multistep_loss}
\end{equation}
where $P_h$ is the hierarchical distribution over valid $h$-step windows.
After an initial single-step training period, separate single-step and
multistep optimizer updates are interleaved.

\subsection{Policy Learning and Deployment}
\label{sec:method_policy}

A navigation policy is trained using a soft actor-critic
(SAC)-based implementation~\cite{haarnoja2018sac}, with both
execution-fitted models fixed.

Conditioning the policy on recent velocity observations allows its decisions
to account for motion responses shaped by preceding commands.
Using the navigation quantities defined in Sec.~\ref{sec:preliminaries},
the policy input is
\begin{equation}
 o_t=[\ell_t^\top,d_t,\beta_t,s_t^\top,
                   s_{t-1}^\top,\ldots,s_{t-H_s}^\top]^\top.
 \label{eq:method_observation}
\end{equation}

Here $H_s$ is the number of preceding velocity estimates, ordered from
most recent to oldest. Unlike the pose model's command history, this
history represents recent execution response rather than control inputs.
For LiDAR ranges, the policy encoder applies the IPAPRec reciprocal
transform~\cite{zhang2022ipaprec} with a learnable offset,
emphasizing nearby obstacles while compressing distant-range variation.

At each step, the policy action is mapped to a physical command by
Eq.~\eqref{eq:method_action}. The execution-fitted simulator returns the next
observation and the reward in Eq.~\eqref{eq:method_reward}. SAC optimizes
the entropy-regularized expected discounted return.
Actor and critic updates use replayed transitions, without propagating
policy gradients through the simulator. Batched model inference supports
parallel training environments.

At deployment, the policy input comprises live LiDAR, localized target
geometry, and a short history of velocity estimates. Because the
observation model is fitted to recorded velocity estimates, live
estimator outputs replace its predictions while preserving the meaning
of the velocity feedback. These live estimates update the policy history,
which is reset at episode boundaries. The robot executes policy commands
through its existing low-level controller. The resulting motion and sensor
feedback replace the execution-fitted simulator; neither fitted model runs
online.

\section{Experiments}
\label{sec:experiments}

\subsection{Data Collection and Training}
\label{sec:experimental-setup}

Policy training was conducted on an NVIDIA GeForce RTX 5090 using
FlashNav~\cite{wang2026flashnav}, a GPU-parallel two-dimensional simulator for
efficient navigation-policy learning.
Dynamics models were fitted separately to each platform's execution data and
used to train its corresponding policies. Both platforms received navigation
commands at 10\,Hz.
Nominal denotes the baseline configuration, in which a navigation policy
was trained using the simulator's predefined motion model and current
velocity observations, without the proposed execution-fitted dynamics or
velocity-history input. The predefined model integrates simulated
velocities to update robot pose and supplies those velocities as policy
feedback. Prediction experiments compare this motion model with the
execution-fitted model, while navigation experiments compare the resulting
policies. The ablation study independently varies the motion model, policy
history, and feedback source.

Identification data comprised velocity commands, reference poses,
and body-frame velocity estimates. Unitree Go2 reference poses were obtained from
FAST-LIO2~\cite{xu2022fastlio2}, with velocity feedback provided through
Unitree's Sport state interface. Jackal used reference poses from the target
simulator and filtered-odometry velocity estimates. Training windows were constructed
from valid, continuous sequences.

For both platforms, the datasets combined steady-state excitation data with
recordings from policy-driven navigation and were split into training and
validation sets by trajectory group. The two data sources received training weights of
20\% and 80\%, respectively. Go2 provided 26,495 training samples and
19,842 validation samples; Jackal provided 32,225 training samples and
6,401 validation samples. Each sample represents one control interval.

Pose prediction used an MLP with two 256-unit hidden layers, sigmoid linear unit
(SiLU) activations, and a three-dimensional linear output. Excluding the
interval duration, the state--command input comprised 21 features for Go2 and
14 for Jackal.
An independent ARX model was fitted by weighted ridge regression with
a regularization coefficient of $10^{-3}$. The checkpoint was selected by
single-step normalized mean squared error (MSE) on the grouped validation set.

Policy training used four 128-unit hidden layers for both the actor and critic,
a batch size of 512, a discount factor of 0.99, and a learning rate of
3\,\texttimes\,10\textsuperscript{\textminus4}. Across three random seeds, training for two million environment transitions
required 4.02\,\textpm\,0.03\,min for Jackal and 4.97\,\textpm\,0.01\,min for Go2.
After training, the selected policy checkpoints were frozen for subsequent
evaluation.
Both Go2 policies evaluated in Scene 1
(Sec.~\ref{sec:real-deployment}) were trained with a
0.75\,\texttimes\,0.36\,m robot footprint.


\subsection{Dynamics Model Validation}
\label{sec:model-validation}
\label{sec:results}

Prediction accuracy was evaluated against the motion models used in the
Nominal configuration over
horizons from 0.1 to 5\,s, using matched continuous validation windows
within each platform. Each rollout was initialized with recorded
velocity feedback and command history, then advanced using recorded
commands and recursively predicted feedback. Position and yaw root-mean-square errors (RMSEs)
were computed at the rollout endpoint, with yaw differences wrapped
to $[-\pi,\pi]$.

\begin{figure}[t]
 \centering
 \includegraphics[width=\columnwidth]{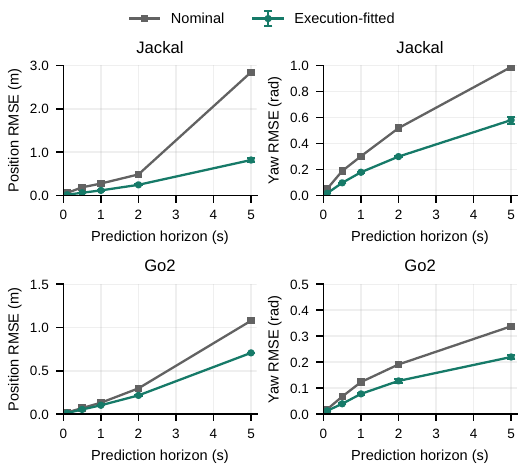}
 \caption{Pose-prediction accuracy on the Jackal and Go2 validation sets.
 Curves show mean endpoint RMSE over the prediction horizon; error bars indicate
 one standard deviation across three pose-model seeds.}
 \label{fig:jackal_rollout}
 \label{fig:platform-rollout}
\end{figure}

Fig.~\ref{fig:platform-rollout} shows lower position and yaw RMSE
for the execution-fitted response at all evaluated horizons on the full
validation sets. At 5\,s, position and yaw RMSE decreased by 71.3\% and
41.3\% for Jackal, and by 34.4\% and 34.9\% for Go2, respectively.
Thus, the improvement was sustained across platforms but was larger for
Jackal over the evaluated horizon. This comparison covers the
identification-data range, not only deployment commands. Within the Go2
deployment command envelope, the execution-fitted model retained lower yaw
error but did not consistently improve position accuracy.




\begin{figure}[t]
 \centering
 \begingroup
 \newlength{\overviewpanelheight}
 \setlength{\overviewpanelheight}{0.27\columnwidth}
 \begin{minipage}[t]{0.29\columnwidth}
  \centering
  \parbox[c][\overviewpanelheight][c]{\linewidth}{\centering
   \includegraphics[width=0.98\linewidth,height=\overviewpanelheight,
   keepaspectratio]{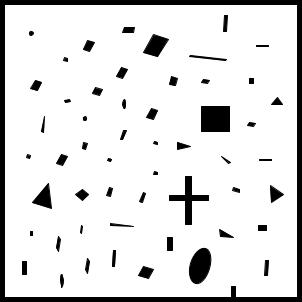}}
  \par\smallskip
  \footnotesize (a)
 \end{minipage}\hfill
 \begin{minipage}[t]{0.38\columnwidth}
  \centering
  \parbox[c][\overviewpanelheight][c]{\linewidth}{\centering
   \includegraphics[width=0.98\linewidth,height=\overviewpanelheight,
   keepaspectratio,trim=44bp 0 44bp 0,clip]{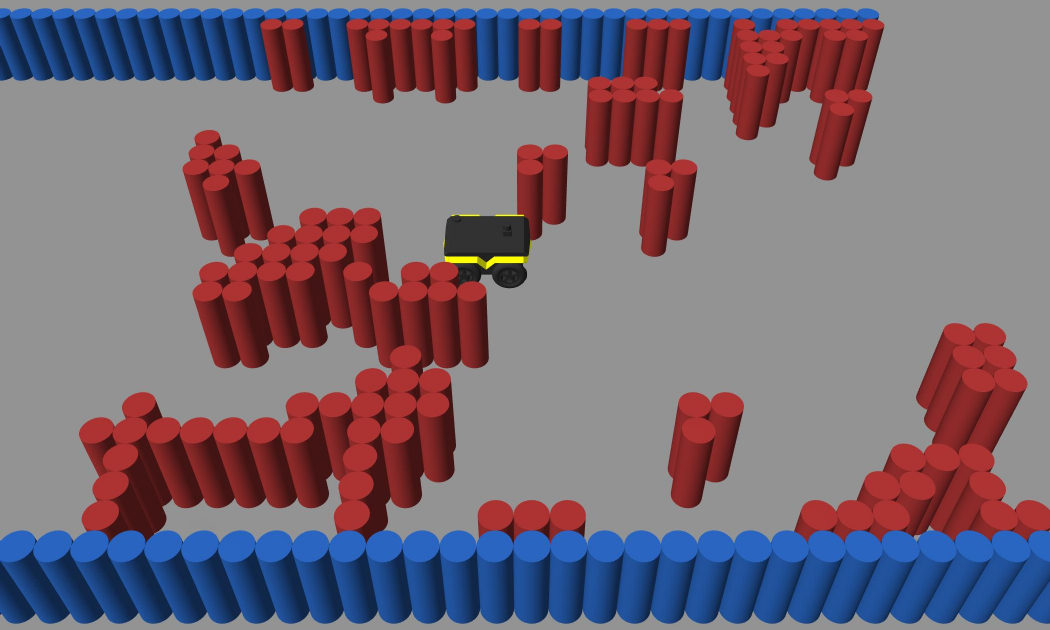}}
  \par\smallskip
  \footnotesize (b)
 \end{minipage}\hfill
 \begin{minipage}[t]{0.29\columnwidth}
  \centering
  \parbox[c][\overviewpanelheight][c]{\linewidth}{\centering
   \includegraphics[width=0.98\linewidth,height=\overviewpanelheight,
   keepaspectratio]{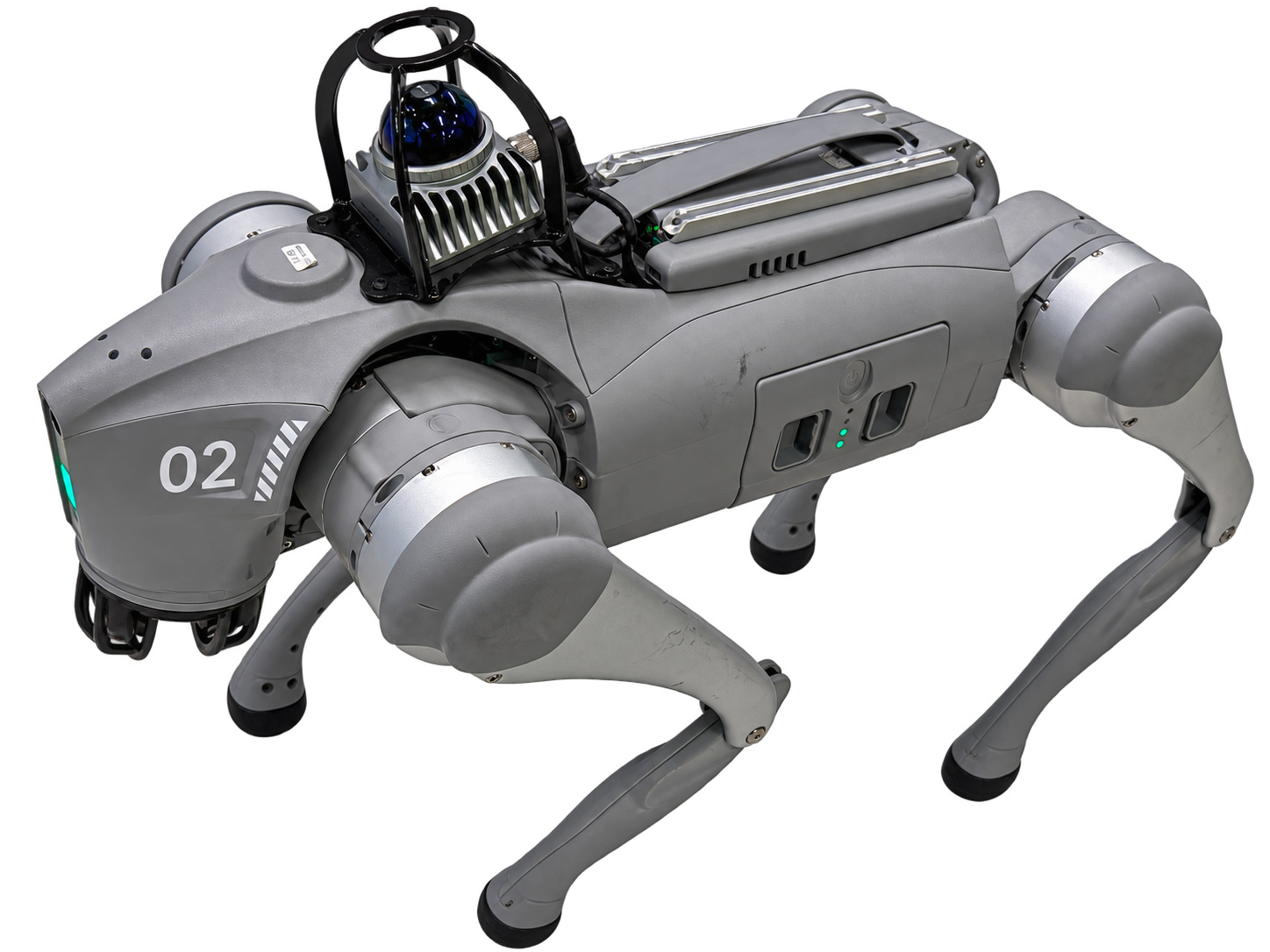}}
  \par\smallskip
  \footnotesize (c)
 \end{minipage}
 \endgroup
 \caption{Training and evaluation settings: (a) Simulation training map,
 (b) Jackal in BARN, and (c) physical Go2.}
 \label{fig:experimental_overview}
\end{figure}

\subsection{Navigation Performance in Simulation}
\label{sec:simulation-experiments}
\begin{table}[t]
\centering
\small
\setlength{\tabcolsep}{2pt}
\renewcommand{\arraystretch}{1.15}
\caption{Simulation-based performance comparison of different models using the BARN dataset.}
\label{tab:external-navigation}
\begin{tabular*}{\columnwidth}{@{\extracolsep{\fill}}cccccc@{}}
\toprule[0.9pt]
Method & SR (\%) $\uparrow$ & CR (\%) $\downarrow$ & TR (\%) $\downarrow$
& Metric $\uparrow$ & Global \\
\hline
DWA & 46.00 & 37.00 & 17.00 & 0.2130 & Yes \\
E-Band & 57.70 & 9.70 & 32.60 & 0.2880 & Yes \\
Nominal & 71.22 $\pm$ 5.68 & 26.89 & 1.89 & 0.3547 & No \\
Nominal & 80.78 $\pm$ 1.35 & 18.44 & 0.78 & 0.3971 & Yes \\
RWM-U & 69.56 $\pm$ 4.40 & 25.56 & 4.89 & 0.3469 & No \\
RWM-U & 73.78 $\pm$ 5.67 & 25.56 & \textbf{0.67} & 0.3680 & Yes \\
EIDA & 81.33 $\pm$ 0.58 & 9.33 & 9.33 & 0.3984 & No \\
EIDA & \textbf{89.33} $\boldsymbol{\pm}$ \textbf{2.33} & \textbf{9.22} & 1.44 & \textbf{0.4241} & Yes \\
\bottomrule[0.9pt]
\end{tabular*}
\end{table}

Navigation policies were evaluated on 100 BARN worlds~\cite{perille2020barn}.
Gazebo~\cite{koenig2004gazebo} served as a physics-based target environment for
assessing sim-to-sim transfer across different command-to-motion dynamics.
Each policy was evaluated three times per world, yielding 300 trials
per policy and planner setting. Global paths were generated by Navfn.
Success required reaching within 1.0\,m of the goal without collision
within 100\,s. The robot used a 2D LiDAR with a 270\textdegree\ field of view,
with maximum linear and angular speeds of 2\,m/s and $\pi$/3\,rad/s,
respectively. No additional safety controller or artificial action
delay was introduced.

Simulation tables report success rate (SR), collision rate (CR),
timeout rate (TR), and the mean BARN navigation score. The score
$s_i$ for the $i$th trial is defined as~\cite{xiao2026barnchallenge}
\begin{equation}
 s_i=\mathbf{1}_i^{\text{success}}\times
 \frac{\mathrm{OT}_i}
 {\operatorname{clip}(\mathrm{AT}_i,2\mathrm{OT}_i,8\mathrm{OT}_i)},
 \label{eq:experiment_metric}
\end{equation}
where $\mathbf{1}_i^{\text{success}}$ equals one when the robot reaches
the goal without collision and zero otherwise. $\mathrm{AT}_i$ and
$\mathrm{OT}_i$ denote actual and optimal traversal time, respectively.
Optimal traversal time is computed as
\begin{equation}
 \mathrm{OT}_i=
 \frac{\text{Path Length}_i}{\text{Maximum Speed}},
 \label{eq:experiment_optimal_time}
\end{equation}
where Path Length is the Dijkstra reference path supplied by BARN and
Maximum Speed is 2\,m/s. The clip function bounds $\mathrm{AT}_i$ between
$2\mathrm{OT}_i$ and $8\mathrm{OT}_i$. The reported score is the
mean of $s_i$ over all trials.

We compare EIDA with the dynamic window approach (DWA)~\cite{fox1997dwa},
elastic-band planning (E-Band)~\cite{quinlan1993elastic}, Nominal, and
RWM-U~\cite{li2025rwmu}. We adapted RWM-U to the Jackal command and pose
interfaces and trained its policy using the same SAC algorithm with a nonzero
uncertainty penalty.
The effect of velocity history in the policy input is evaluated separately in
Sec.~\ref{sec:simulation-ablations}. DWA and E-Band use Navfn global paths,
while learned policies are tested both with and without this guidance.

Table~\ref{tab:external-navigation} reports SR as the mean and sample
standard deviation across three policy-training seeds. CR, TR, and the score
are averaged across the same seeds. Classical-planner rows report aggregate
results under the corresponding evaluation protocol.
\begin{figure}[t]
 \centering
 \includegraphics[width=\columnwidth]{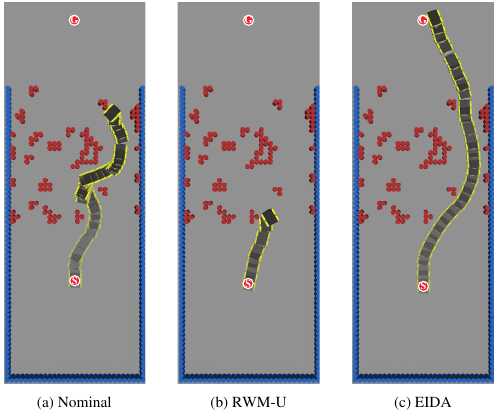}
 \caption{Trajectories of the robot controlled by (a) Nominal, (b) RWM-U,
 and (c) EIDA in one of the 100 BARN testing environments. ``S'' represents
 the starting point, and ``G'' represents the target point. The corresponding
 videos are provided in the supplementary material.}
 \label{fig:barn-trajectories}
\end{figure}

EIDA achieved the highest success rate and score among the learned policies
under both planner settings. Relative to Nominal, the strongest learned
comparator in Table~\ref{tab:external-navigation}, EIDA improved success by
10.11 percentage points without global guidance and by 8.55 points with it.
In the latter setting, its success rate was 89.33\% $\pm$ 2.33\%, with a score
of 0.4241; its margins over RWM-U were larger in both settings. These results
indicate that fitting the target-system execution interface, together with
velocity-history policy input, provided a stronger navigation basis than the
Nominal and recurrent-model baselines under the common evaluation protocol.
As shown in Fig.~\ref{fig:barn-trajectories}, Nominal exhibited lateral drift,
followed by an abrupt heading
correction that caused its rear footprint to sweep into a nearby obstacle.
RWM-U collided shortly after entering the obstacle field. In contrast, EIDA
followed a smooth trajectory through the clutter and reached the goal.
\subsection{Ablation Studies in Simulation}
\label{sec:simulation-ablations}

Table~\ref{tab:barn} reports a complete \(2\times2\times2\) factorial
ablation of the motion model (Nominal or Fitted), the inclusion of velocity
history in the policy input (No or Yes), and the feedback source (Command or
ARX), together with an additional Pose-derived feedback configuration. All
policies receive the current velocity estimate, while the history setting
additionally
provides the five preceding estimates. The pose predictor retains its
command-history input in every Fitted configuration. EIDA denotes the complete
Fitted/Yes/ARX configuration. The additional Pose-derived configuration
retains the fitted pose model and velocity history but computes velocity
feedback from the
predicted pose increment. Each configuration was evaluated on 100 BARN worlds
with three repeats, both with and without global guidance.

\begin{table}[t]
\centering
\small
\setlength{\tabcolsep}{1.2pt}
\renewcommand{\arraystretch}{1.2}
\caption{Ablation of motion modeling, velocity history, and feedback source on BARN.}
\label{tab:barn}
\begin{tabular*}{\columnwidth}{@{\extracolsep{\fill}}ccccccc@{}}
\toprule[0.9pt]
\multirow{2}{*}{Model} & \multirow{2}{*}{Hist.} & \multirow{2}{*}{Feedback} & \multicolumn{2}{c}{Global path: No} & \multicolumn{2}{c}{Global path: Yes} \\
\cline{4-5}\cline{6-7}
& & & SR (\%) & Metric & SR (\%) & Metric \\
\hline
Nominal & No & Command & 71.67 & 0.3555 & 81.00 & 0.3977 \\
Nominal & Yes & Command & 70.67 & 0.3525 & 83.33 & 0.4082 \\
Nominal & No & ARX & 70.67 & 0.3505 & 83.33 & 0.4109 \\
Nominal & Yes & ARX & 72.33 & 0.3611 & 76.00 & 0.3726 \\
Fitted & No & Command & 73.33 & 0.3622 & 82.00 & 0.3954 \\
Fitted & Yes & Command & 76.00 & 0.3667 & 84.00 & 0.3887 \\
Fitted & No & ARX & 73.33 & 0.3612 & 81.33 & 0.3909 \\
Fitted & Yes & Pose-derived & 53.00 & 0.2320 & 46.67 & 0.2079 \\
EIDA & Yes & ARX & \textbf{81.67} & \textbf{0.4025} & \textbf{92.00} & \textbf{0.4374} \\
\bottomrule[0.9pt]
\end{tabular*}
\end{table}

EIDA achieved the highest success rate and score under both planner settings.
The contribution of each component can be assessed by removing one factor at a
time from the complete configuration. Replacing the fitted motion model with
Nominal reduced success by 9.34 and 16.00 percentage points without and with
global guidance, respectively. Removing velocity history reduced success by
8.34 and 10.67 points, while replacing ARX feedback with the command-driven
simulator state reduced it by 5.67 and 8.00 points. All three removals also
reduced the navigation score. These controlled comparisons show that motion
fitting, velocity history, and ARX feedback each make a distinct
contribution to the complete EIDA configuration.

The contribution of velocity history is consistent with the temporal
dependence represented by the fitted dynamics model. Because the pose
predictor conditions on recent commands, similar current commands and velocity
estimates can lead to different subsequent motion depending on the preceding
execution sequence. Velocity history provides the policy with information
about this recent response. Performance changes across the partial
configurations are not uniformly additive, suggesting that the effect of each
component depends on the surrounding interface configuration.

The feedback-source comparisons further support the need to reproduce the
policy-facing velocity signal. With the fitted pose model and velocity history
fixed, ARX feedback outperformed both the command-driven state and Pose-derived
feedback.
Relative to ARX, Pose-derived feedback reduced success by 28.67 percentage
points without global guidance and by 45.33 points with it; the score in the
latter setting decreased by 52.5\%. Pose-derived feedback represents an average
velocity computed from the predicted displacement over one control interval,
whereas ARX directly models the velocity estimate available to the deployed
policy. These comparisons support modeling feedback separately from the
geometric state update.

\begin{figure}[t]
 \centering
 \begin{minipage}[t]{0.485\columnwidth}
  \centering
  \includegraphics[width=\linewidth]{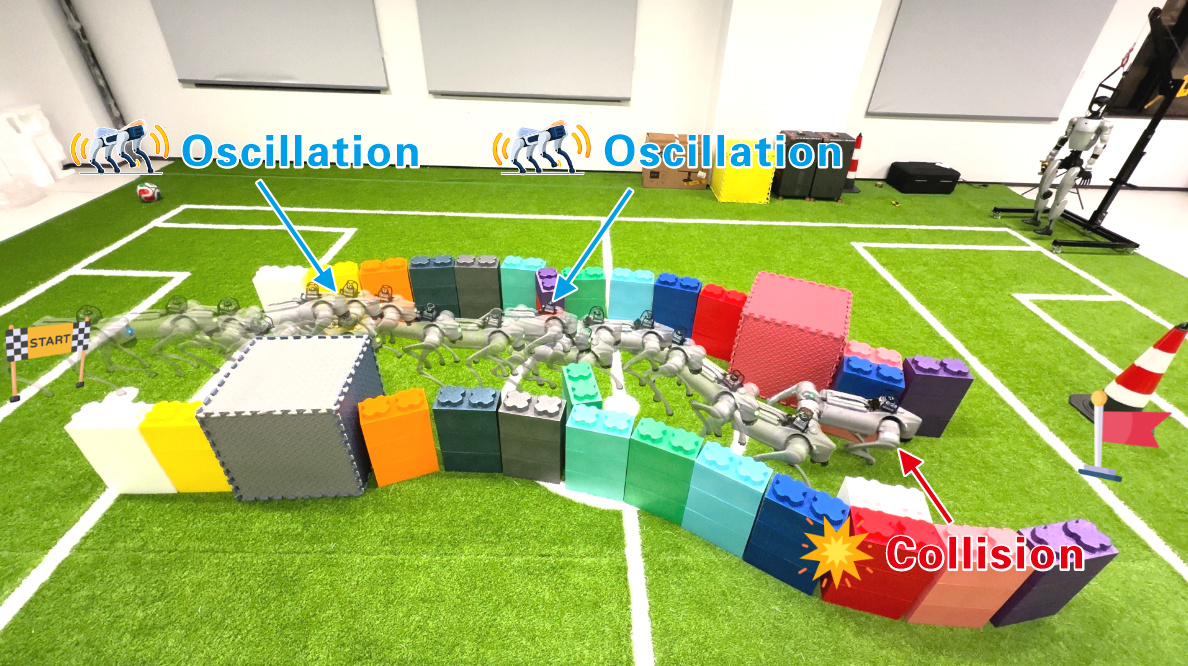}\par
  \vspace{2pt}
  {\footnotesize (a) Scene 1: Nominal\par}
 \end{minipage}\hfill
 \begin{minipage}[t]{0.485\columnwidth}
  \centering
  \includegraphics[width=\linewidth]{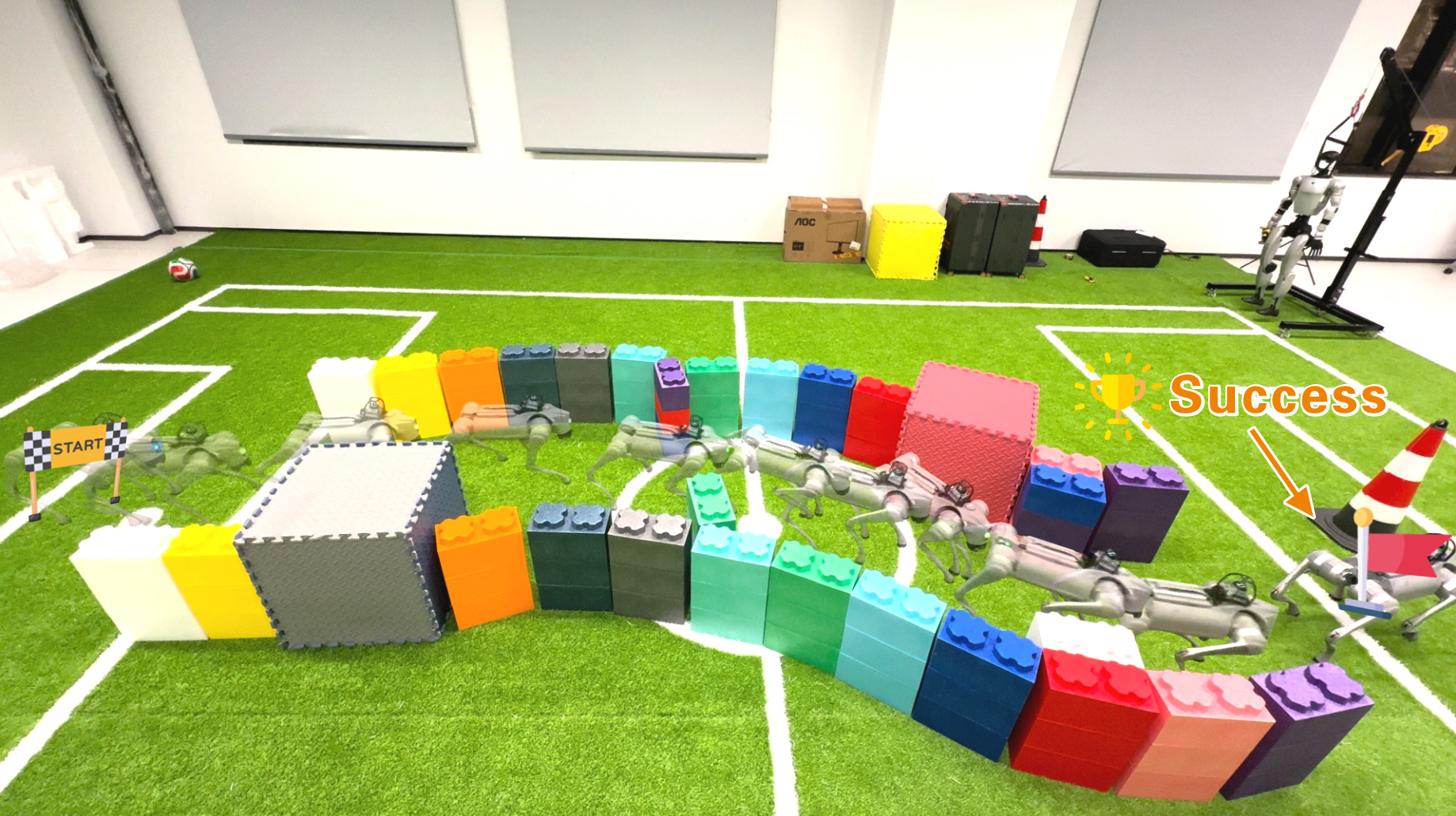}\par
  \vspace{2pt}
  {\footnotesize (b) Scene 1: EIDA\par}
 \end{minipage}\par
 \vspace{5pt}
 \begin{minipage}[t]{0.485\columnwidth}
  \centering
  \includegraphics[width=\linewidth]{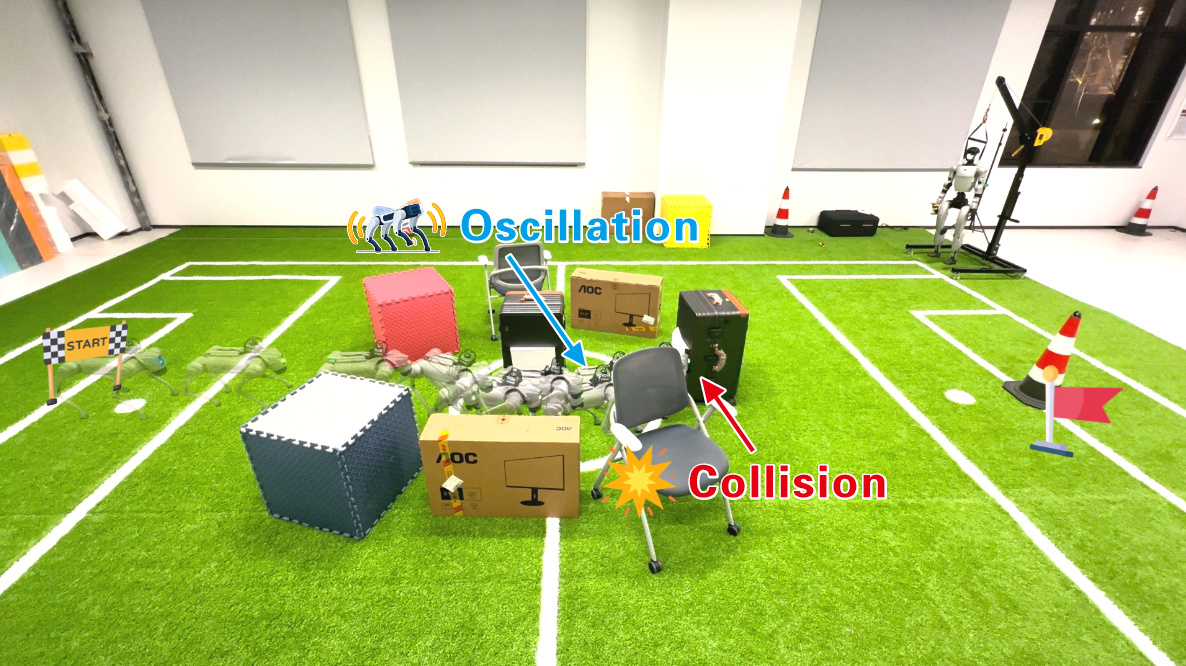}\par
  \vspace{2pt}
  {\footnotesize (c) Scene 2: Nominal\par}
 \end{minipage}\hfill
 \begin{minipage}[t]{0.485\columnwidth}
  \centering
  \includegraphics[width=\linewidth]{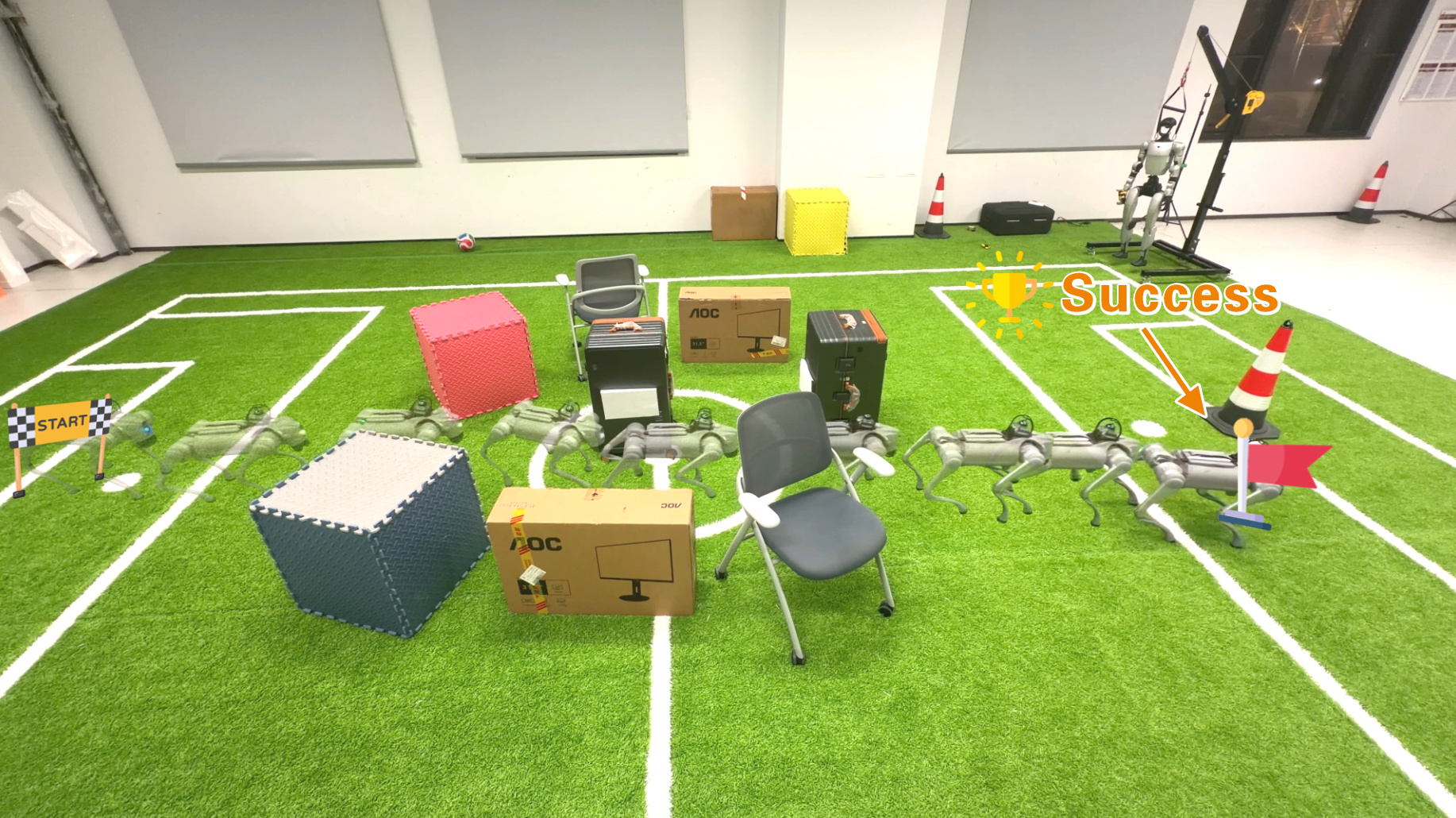}\par
  \vspace{2pt}
  {\footnotesize (d) Scene 2: EIDA\par}
 \end{minipage}
 \caption{Physical Go2 navigation in two static scenes. Panels (a)--(d)
 compare the Nominal baseline with EIDA; a dynamic scenario in which a
 pedestrian suddenly obstructs the path is shown separately in
 Fig.~\ref{fig:go2_dynamic}.}
 \label{fig:go2_static_scenes}
\end{figure}

\begin{figure}[t]
 \centering
 \includegraphics[width=\columnwidth]{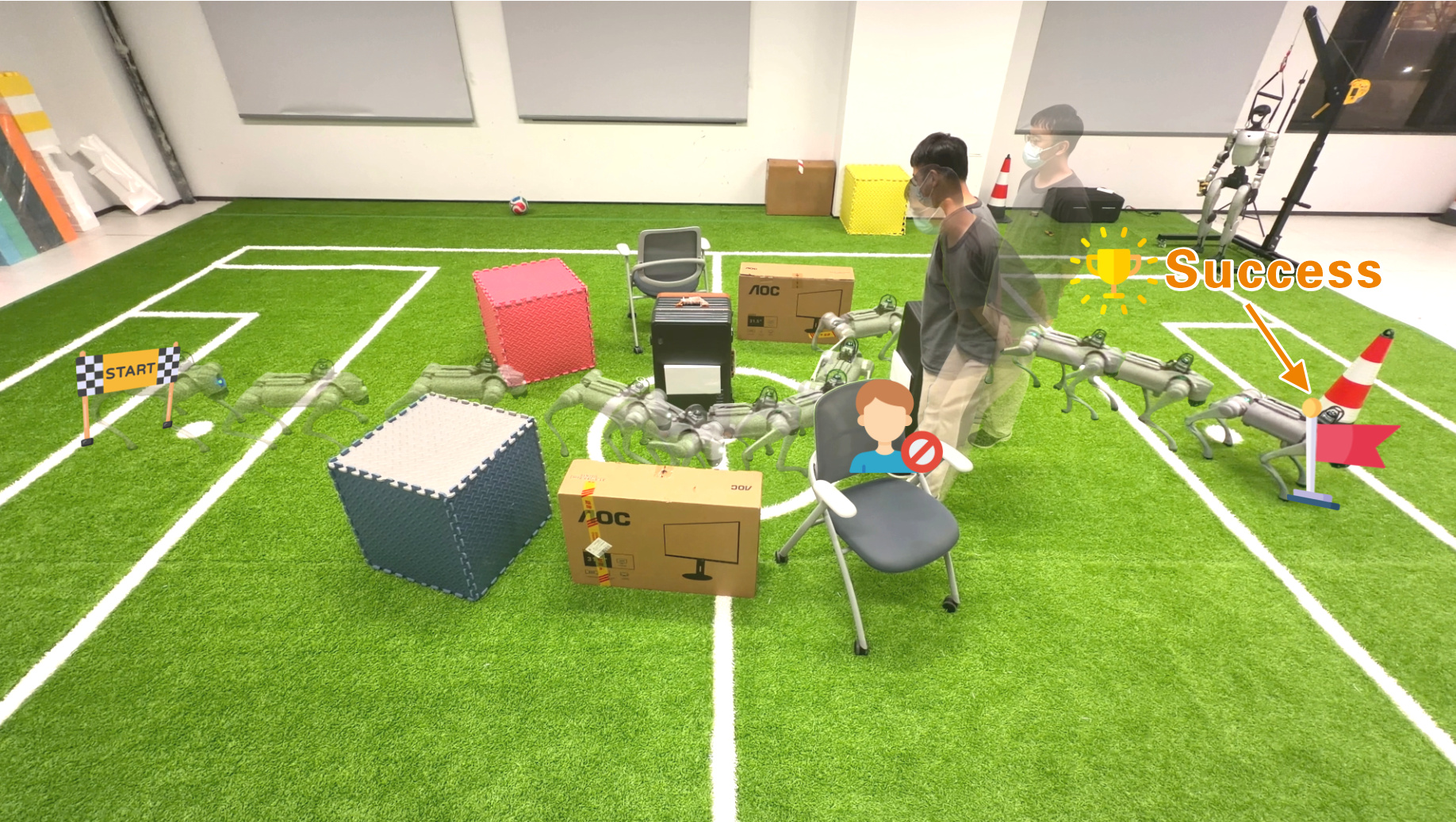}
 \caption{Go2 navigation with EIDA in Scene 2 when a pedestrian suddenly
 obstructs the path. Overlaid robot and pedestrian poses illustrate the
 motion sequence toward the marked goal. This demonstration is separate
 from the static-scene trials in Table~\ref{tab:real_go2}.}
 \label{fig:go2_dynamic}
\end{figure}

\subsection{Real-World Deployment}
\label{sec:real-deployment}

\subsubsection*{a. Platform and Deployment Setup}
Real-world experiments were conducted on a Unitree Go2 quadruped
equipped with a Livox MID-360 LiDAR. The navigation policy ran on an
onboard NVIDIA Jetson Orin NX module with 16\,GB of memory.
FAST-LIO2~\cite{xu2022fastlio2} provided robot pose estimates for
computing relative goal geometry, while body-frame velocity estimates
were obtained through the Sport state interface. The policy issued
velocity commands at 10\,Hz through the robot's existing low-level
control interface.
Commands were limited to $v_x\in[0,1]$\,m/s,
$v_y\in[-0.2,0.2]$\,m/s, and $\omega\in[-\pi/3,\pi/3]$\,rad/s.
Nominal and EIDA used current velocity observations and current
plus five past velocity observations, respectively. For Go2, the baseline training
simulator used a 0.1\,s time step, with forward, lateral, and angular
acceleration limits of 2.0\,m/s\textsuperscript{2},
1.0\,m/s\textsuperscript{2}, and $\pi$\,rad/s\textsuperscript{2}, respectively.

\begin{table}[t]
\centering
\small
\setlength{\tabcolsep}{1.5pt}
\renewcommand{\arraystretch}{1.15}
\caption{Real-world Go2 navigation performance in two scenes.}
\label{tab:real_go2}
\begin{tabular*}{\columnwidth}{@{\extracolsep{\fill}}ccccc@{}}
\toprule[0.9pt]
Policy & Collision $\downarrow$ & Goal $\uparrow$ & Path (m) $\downarrow$ & Time (s) $\downarrow$ \\
\hline
\multicolumn{5}{@{}l}{\textit{Scene 1: Curved passage}} \\
Nominal & 9/10 & 1/10 & 6.81 & 12.93 \\
EIDA & \textbf{0/10} & \textbf{10/10} & \textbf{6.32} $\boldsymbol{\pm}$ \textbf{0.18} & \textbf{10.31} $\boldsymbol{\pm}$ \textbf{0.75} \\
\multicolumn{5}{@{}l}{\textit{Scene 2: Separated obstacles}} \\
Nominal & 7/10 & 3/10 & 6.71 $\pm$ 0.37 & 12.04 $\pm$ 2.29 \\
EIDA & \textbf{0/10} & \textbf{10/10} & \textbf{5.91} $\boldsymbol{\pm}$ \textbf{0.30} & \textbf{8.98} $\boldsymbol{\pm}$ \textbf{1.04} \\
\bottomrule[0.9pt]
\end{tabular*}
\end{table}

\subsubsection*{b. Scenarios and Evaluation Protocol}
Evaluation used two indoor artificial-turf scenes with
different obstacle layouts, as shown in Fig.~\ref{fig:go2_static_scenes}.
Scene 1 contained a curved passage requiring successive heading changes,
whereas Scene 2 contained separated obstacle clusters with gaps between
them. In each trial, the robot navigated from a marked start position
to a designated goal. These layouts tested its ability to adjust its
motion while maintaining clearance from nearby obstacles.

Each policy was evaluated in
ten trials per scene, giving 40 trials in total. The goal radius and
trial time limit were set to 0.3\,m and 30\,s, respectively.
Goal denotes reaching the target without any collision, with outcomes
verified by manual review. Path length and
execution time were summarized only over these successful trials as means
and sample standard deviations; the standard deviation was omitted when
only one trial qualified. Path length was computed from accumulated planar
pose increments during execution, and execution time covered the interval
from the first acknowledged nonzero velocity command to the first
acknowledged stop command after the final nonzero command.

\subsubsection*{c. Experimental Results}
Fig.~\ref{fig:go2_static_scenes}(a)--(d) illustrates the differences
in navigation behavior. The illustrated Nominal trials exhibit oscillatory
motion and contact with nearby obstacles, whereas EIDA
traverses the curved passage and the gaps between obstacle clusters without
collision. The two rows correspond to the same scene identities used in
Table~\ref{tab:real_go2}.

Table~\ref{tab:real_go2} summarizes the corresponding quantitative results.
Collision denotes trials with at least one collision.
EIDA reached the goal without collision in all 20 trials, compared with
4 of 20 trials for the Nominal baseline. Among collision-free goal-reaching
trials, EIDA reduced mean path length by 7.2--11.9\% and execution time by
20.2--25.4\% across the two scenes. These results show more reliable goal
reaching, with shorter paths and lower execution times among successful trials.

In the dynamic scenario shown in Fig.~\ref{fig:go2_dynamic}, a pedestrian
suddenly obstructed the path. EIDA adjusted its motion as the available free
space changed, maintained clearance from the pedestrian, and subsequently
reached the goal. The resulting sequence indicates that the
execution-adapted policy retained stable local navigation behavior under
dynamic interference.

\section{Conclusion}
\label{sec:conclusion}

This work presented an execution-interface dynamics adaptation framework for
transferring navigation policies from lightweight simulation to target
robots. Model validation, controlled BARN navigation, and physical Go2
experiments showed the benefits of learning target-system execution responses
for policy training. Across both real-world scenes, EIDA reached
the goal without collision in all 20 trials and achieved shorter paths and
lower execution times than the Nominal baseline among collision-free
goal-reaching trials.
The framework enables platform-specific adaptation through an existing
velocity-command interface without reconstructing low-level actuation
dynamics. Future work will address long-horizon prediction and
uncertainty-aware modeling.


\ifpaperreferences
\bibliographystyle{IEEEtran}
\bibliography{references}
\fi

\end{document}